# Harvesting AlphaEarth: Benchmarking the Geospatial Foundation Model for Agricultural Downstream Tasks

Yuchi Ma[1*], Yawen Shen[2], Anu Swatantran[2], David B. Lobell[1]

[1]Department of Earth System Science and Center on Food Security and the Environment, Stanford University, USA
[2]Corteva Agriscience™, USA

*Correspondence to*: Yuchi Ma (yuchima8@gmail.com)

**Keywords**: Geospatial Foundation Model, AlphaEarth Foundation, Yield Prediction, Agricultural Practice Mapping

**Abstract**

Geospatial foundation models (GFMs), pretrained on massive Earth observations (EO), have emerged as a promising approach to overcoming the limitations in existing featurization methods. Although most studies on GFMs have released the source codes and pre-trained weights, their deployment still demands extensive configuration, environment setup, inference EO preparation, and model fine-tuning. More recently, Google DeepMind has introduced AlphaEarth Foundation (AEF), a GFM pre-trained using multi-source EOs across continuous time. An annual and global embedding dataset is produced using AEF that is ready for analysis and modeling. The internal experiments show that AEF embeddings have outperformed operational models in 15 EO tasks without re-training. However, those experiments are mostly about land cover and land use classification. Applying AEF and other GFMs to agricultural monitoring require an in-depth evaluation in critical agricultural downstream tasks. There is also a lack of comprehensive comparison between the AEF-based models and traditional remote sensing (RS)-based models under different scenarios, which could offer valuable guidance for researchers and practitioners. This study addresses some of these gaps by evaluating AEF embeddings in three agricultural downstream tasks in the U.S., including crop yield prediction, tillage mapping, and cover crop mapping. Datasets are compiled from both public and private sources to comprehensively evaluate AEF embeddings across tasks at different scales and locations, and RS-based models are trained as comparison models. AEF-based models generally exhibit strong performance on all tasks and are competitive with purpose-built RS-based models in yield prediction and county-level tillage mapping when trained on local data. However, we also find several limitations in current AEF embeddings, such as limited spatial transferability compared to RS-based models, low interpretability, and limited time sensitivity. These limitations recommend caution when applying AEF embeddings in agriculture, where time sensitivity, generalizability, and interpretability is important. To our knowledge, this is the first study that systematically implements and evaluates embeddings from GFMs in agricultural downstream tasks across space, time, and spatial resolutions. The evaluation results and analyses can inform the design of future AEF versions and other GFMs and support their applications in agriculture and Earth science domains. Moreover, the proposed benchmarking workflow and datasets can be readily applied to evaluate future GFMs and facilitate their use in agricultural downstream applications. The code and data will be publicly available at https://github.com/yuchima8/Harvest_AlphaEarth.

1. Introduction

Earth observation (EO), grounded in remote sensing imagery, has enabled scalable and timely monitoring of dynamic Earth system processes, with agriculture being one of its most widely applied domains. Since the launch of the first Landsat satellite in 1972, there were early studies that explored the association between satellite spectral observations and crop conditions on the ground (Deering, 1978; Kauth and Thomas, 1976). Over the past half century, the rapid expansion of satellite platforms and the availability of Earth observations have profoundly transformed research and practices in agroecosystems, such as cropland mapping (Zhang et al., 2025), crop yield prediction (Xiong et al., 2026), agricultural practice detection (Luo et al., 2023), agricultural carbon outcome measurement (Guan et al., 2023), and food security assessment (Nakalembe et al., 2021).

Despite the progress, there are still several bottlenecks that limit the application of EO in agriculture. A major challenge lies in the varying availability of EO data, driven by differences in satellite track overlap and cloud cover. Infrequent observations or data gaps can substantially reduce the effectiveness of cropland monitoring. Data harmonization is another common challenge when multi-modal EO data are used, arising from discrepancies in sensor design, spatial resolution, spectral response functions, and temporal revisit frequencies. In the past decade, machine learning (ML) and deep learning (DL) models based on EO have demonstrated potential across multiple agricultural downstream tasks. However, training ML/DL models requires abundant, high-quality, and task-specific labels. These ground truth labels, such as cropland types or crop yield records, are typically collected through expert annotation or field surveys, which are both time-consuming and resource intensive.

Recently, geospatial foundation models (GFMs) have emerged as a promising approach to overcome the abovementioned bottlenecks, as they are trained on broad, large-scale data and can be readily adapted to a wide range of downstream EO tasks. GFMs are typically pre-trained in a self-supervised way, wherein the model exploits the input data itself to construct learning objectives, such as image reconstruction (He et al., 2022) or contrastive learning (Wang et al., 2022). This pre-training process requires no task-specific labels and enables the model to extract generic features from the input remote sensing data that can be adapted to diverse downstream tasks through fine-tuning or zero-shot learning.

Early GFMs were trained based on a single modality and targeted at a few related downstream tasks. For example, SatMAE (Cong et al., 2022), one of the first GFMs, was pre-trained on multi-spectral Sentinel-2 images using Masked Autoencoders (He et al., 2022) and evaluated on cropland classification and semantic segmentation. Similarly, Scale-MAE (Reed et al., 2023) was pre-trained by jointly learning representations from optical images at both low and high scales, achieving improved accuracy in 8 land-use classification datasets. In addition to encoding spatial information, NASA and IBM released a multi-temporal GFM named Prithvi-EO-2.0 (Szwarcman et al., 2024), which explicitly leverages transformer attention across both spatial and temporal dimensions via pre-training on a decade of satellite imagery from the Harmonized Landsat–Sentinel-2 archive. Recent efforts have shifted towards building multi-modal GFMs that accept inputs from more than one source, such as optical, multispectral, hyperspectral, and SAR images. For instance, the DOFA model (Xiong et al., 2024) incorporates a universal feature learning module designed for heterogeneous data modalities and was jointly trained using observations from five sensors. SkySense++ (Wu et al., 2025) integrates data from 11 satellite platforms and was pretrained progressively to learn both general representations and semantically enriched representations.

Although an increasing number of GFMs has been developed and released, several limitations remain in their application to practical downstream tasks. Even though most GFM studies have released source codes and pre-trained models, their deployment still demands extensive configuration and environment setup, posing challenges for users with limited expertise in deep learning or foundation models. Meanwhile, adapting these GFMs requires fine-tuning with task-specific labels together with remote sensing observations, which still involve intensive image collection and pre-processing.

Most recently, Google DeepMind introduced AlphaEarth Foundation (AEF) (Brown et al., 2025), a GFM with ~480M parameters that generates near-global geospatial representation that assimilates spatial, temporal, climate, topography, and measurement contexts across multiple sources. AEF achieves continuous time-series EO featurization via several innovations, such as a Space-Time Precision encoder, an adaptive decoding scheme, and a spatially dense information time-bottleneck. Pre-trained using ~3 billion images sampled from ~5 million locations worldwide, AEF learns general representations that capture temporal dynamics of Earth's surface and climatic activities. At inference, it ingests annual multi-source EOs and encodes Earth's surface into 64-dimentional embeddings at 10 m spatial resolution. The annual AEF embeddings are publicly available on Google Earth Engine in 2017–2024.

AEF can potentially transform EO tasks since it has largely addressed the challenges of data harmonization and data gaps. Internal evaluation showed that AEF embeddings have consistently outperformed previous featurization approaches tested on 15 EO tasks without re-training (Brown et al., 2025). However, certain aspects of AEF remain insufficiently evaluated. First, AEF embeddings have been evaluated mostly on land use and land cover mapping tasks, with no insights into critical agricultural downstream tasks, such as crop yield prediction and agricultural practice mapping. Second, internal evaluations were conducted in the same region and the same period, with no testing of the spatial or temporal generalizability of the AEF embeddings. More importantly, it remains unclear how AEF embeddings compare with commonly used remote sensing features across different scenarios, an evaluation that could offer valuable guidance for researchers and practitioners.

To answer these questions and facilitate better application of AEF embeddings in agriculture and potentially other EO tasks, we proposed a benchmarking workflow that systematically evaluates AEF embeddings across three agricultural downstream tasks: crop yield prediction, tillage classification, and cover crop mapping. Ground truth labels were compiled from both public and private sources to comprehensively evaluate AEF embeddings across tasks at different scales and locations. In addition to local experiments, we also evaluated AEF's generalizability across space, time, and scales. Meanwhile, remote sensing-based models were trained as baselines for comparison in each task.

The structure of the paper is as follows: In Section 2, we describe each of the downstream task and the corresponding sources of the ground truth data. In Section 3, we introduce the AEF embeddings and the remote sensing features. The details about the data downloading and pre-processing procedures are provided. In addition, ML models and evaluation schemes are illustrated. In Section 4, we present the evaluation results of the AEF-based models and RS-based models across three tasks. The space- and scale-transfer experiments have also been conducted. In Section 5, we summarize the advantages and limitations of AEF embeddings in agricultural applications. In Section 6, we conclude this study and discuss potential future research.

## 2. Experimental tasks and ground truth data

We evaluated the potential of AEF embeddings for crop yield prediction, tillage classification, and cover crop mapping at both the regional scale (county level) and the field scale. The corresponding agricultural data were collected as ground truth labels for model training and evaluation. In particular, the county-level agricultural data are provided by the governments' agricultural agencies, which are public available. Field-level agricultural data are typically stored in governmental databases or commercial systems and are hard to acquire due to privacy concerns (Deines et al., 2021; Ma et al., 2024b). Corteva Agriscience collected and provided the agricultural data at the field level in the U.S. for model training and evaluation in this study. More details are given below.

## 2.1 Crop yield prediction

Crop yield prediction provides essential information for farm resource management, food security monitoring, and market planning (Becker-Reshef et al., 2019). Many governments conduct monthly and annual agricultural surveys to estimate crop progress and predict yield within the growing season. More recently, ML-based yield prediction models using EO have proven an effective and cost-efficient approach that can accurately predict crop yields. Specifically, raw remote sensing or derived vegetation indices are used as predictor variables to train ML models, together with the corresponding ground truth yield records. The trained ML models are implemented to predict yields in unseen regions or years.

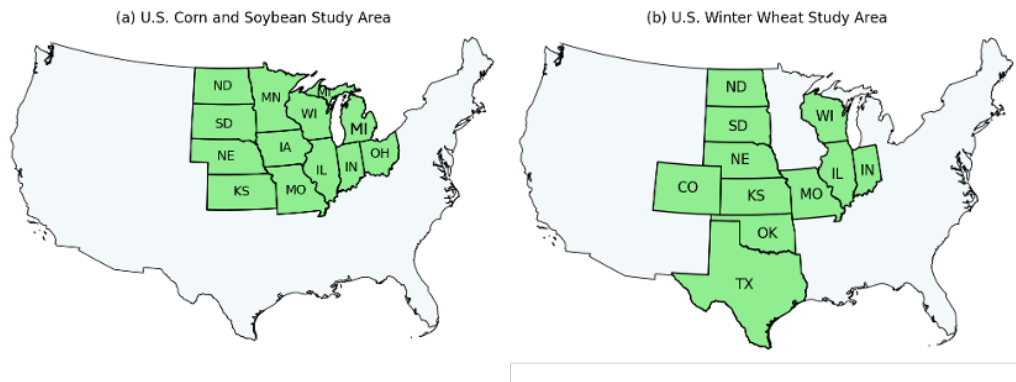

Figure 1 The study areas in the U.S., covering (a) 12 states in the U.S. Corn belt for corn and soybean, and (b) 11 Midwestern states for winter wheat. The states include Colorado (CO), Illinois (IL), Indiana (IN), Iowa (IA), Kansas (KS), Michigan (MI), Minnesota (MN), Missouri (MO), Nebraska (NE), North Dakota (ND), Oklahoma (OK), Texas (TX), South Dekota (SD), and Wisconsin (WI).

We evaluated the performance of AEF embeddings for yield prediction in the U.S., which is the world's largest agricultural producer and export (Kogan et al., 2013). The county-level yield data for corn, soybean, and winter wheat in the U.S. from 2017 to 2024 were obtained from the United States Department of Agriculture National Agricultural Statistics Service (USDA-NASS) (USDA-NASS, 2024), which publishes annual yield statistics after the growing season. Corn and soybean yield data were collected from 12 U.S. Corn Belt states and winter wheat yield data from 11 Midwestern states, which are the major crop producers and account for more than 70% of the national crop production (Figure 1).

In addition, we also evaluated AEF embeddings on field-level yield prediction, which provides fine-scale information to support farm management. The ground-based measures of yields were collected using in-field yield monitoring system during harvesting by Corteva Agriscience (Deines et al., 2021). This Corteva yield datasets include multi-year field-level yield data and the field locations for corn, soybean, and winter wheat. Corn and soybean fields distribute across the U.S. Corn Belt and winter wheat fields span the U.S. Wheat Belt and Midwest (Ma et al., 2024b), including 89,938 corn fields and 73,492 soybean fields in 2017–2018, and 20,401 winter wheat fields in 2017–2022.

## 2.2 Spring tillage mapping

Tillage is an agricultural practice that farmers implement to prepare the land for sowing. Conventional high-intensity tillage often leaves little crop residue and causes large disturbance to the soil, which accelerates soil erosion and increase greenhouse gas emissions (Lu et al., 2022). Low-intensity tillage, including conservation tillage and no-till, has been promoted worldwide to conserve soil and water (Claassen et al., 2017). Due to its far-reaching environmental impacts, tillage mapping based on ML and remote sensing observations has drawn considerable research attention. In addition to raw spectral bands and VIs, tillage-related indices have been derived to detect crop residue cover and classify tillage practices, such as Normalized Difference Tillage Index (NDTI) (Van Deventer et al., 1997).

We evaluated the performance of AEF embeddings for spring tillage mapping in the U.S. Corn Belt at both the county level and the field level (Figure 1 (a)). County-level tillage statistics were collected from USDA-NASS, which record the area of cropland managed under different tillage systems, including conventional tillage, conservation tillage, and no-till (Zulauf and Brown, 2019). The tillage statistics are compiled through the USDA Census of Agriculture every five years and are currently available in 2017 and 2022. We collected the data and combined conservation tillage and no-till as low-intensity tillage. The proportion of cropland under low-intensity tillage within each county were calculated as the ground truth data, and there are overall 2,001 county-year records.

Field-level Spring tillage data were collected from farmers by Corteva Agriscience. The Corteva tillage dataset contains grower-submitted records of tillage operations and the field boundaries spanning 2016–2023 throughout the U.S. Corn Belt. We used the data in 2017–2023 and categorized the tillage activity for each field into low- or high-intensity tillage according to the corresponding intensity levels, resulting in a total of 24,514 field-yield records.

## 2.3 Cover crop mapping

Cover cropping is another conservation agricultural practice in which non-cash crops are cultivated during the interval between the harvest of cash crops and the planting of the next (Plastina et al., 2020). It helps prevent periods of bare soil in croplands and reduce the risk of soil erosion (Koudahe et al., 2022), nitrogen leaching (Abdalla et al., 2019), and weed infestation (Alonso-Ayuso et al., 2018). In the U.S., the adoption of cover crops has expanded substantially in recent decade, with the national cover crop area nearly doubling from 10.3 million acres in 2012 to 18.0 million acres in 2022 (Lobell et al., 2025; Zulauf et al., 2024). EO-based methods have been developed to detect and track the adoption of cover crops (Fendrich et al., 2023; Zhou et al., 2022).

We evaluated the performance of AEF embeddings for cover crop mapping in the U.S. Corn Belt at the field level (Figure 1 (a)). Again, the field-level cover crop data were collected from farmers by Corteva Agriscience. Each data record specified if a cover crop was used in a field and provided the field boundary, the start and end dates of cover cropping, and the species planted. The annual cover crop dataset covers the 12 states in the U.S. Corn Belt and spans from 2017 to 2024. Since the planting dates of the cover crops can happen in the fall after the harvest, or in the spring before the planting, we concatenated two-year AEF embeddings as the input predictors. Therefore, we dropped year 2017 and did experiments for 2018–2024, leading to a total number of 47,709 field-year samples. We did not conduct county-level cover crop mapping experiments because county-level data are available only for 2017 and 2022, and AEF embeddings for 2016 are unavailable, which prevents the construction of two-year AEF embedding inputs as predictors.

A summary of the datasets and experimental settings for each downstream tasks is given in Table 1.

Table 1 A summary of the experimental settings.

| Task | Level | Year | Crop | # Samples |
|---|---|---|---|---|
| Yield Prediction | County | 2017–2024 | Corn | 6,325 |
| | | | Soybean | 6,024 |
| | | | Winter Wheat | 3,020 |
| | Field | 2017–2018 | Corn | 89,938 |
| | | | Soybean | 73,492 |
| | | 2017–2022 | Winter Wheat | 20,401 |
| Tillage Mapping | County | 2017&2022 | N/A | 2,001 |
| | Field | 2017–2023 | N/A | 24,514 |
| Cover Crop Mapping | Field | 2018–2024 | N/A | 47,709 |

## 3. Materials

### 3.1 Alpha Earth Foundation model

The AEF model was trained on ~3 billions of observations across optical (Landsat 8/9, Sentinel-2), radar (Sentinel-1), LiDAR (GEDI), Climate (ERA5-Land), gravity fields (GRACE), Elevation (GLO-30), and text sources (Wikipedia) (Brown et al., 2025). A space–time encoder and a teacher–student framework was employed to capture spatial, temporal, and measurement contexts in a compact form. Consequently, it transformed sparse, heterogeneous EOs from multiple sources into a universal 64-dimensional embedding space for global mapping. The resulting 64-dimensional embeddings are time-continuous and sensor-agnostic at 10-meter resolution, achieving near-global spatial coverage and excluding polar extremes beyond ~82° N/S.

The AEF embeddings are publicly available as annualized, global layers spanning 2017 to 2024, hosted on Google Earth Engine (GEE) as an image dataset. The 64-band annual embeddings are stored as image layers, with individual bands labeled sequentially from A00 to A63.

### 3.2 Remote sensing, climate, and topographic data

In addition to AEF embeddings, we extracted traditionally used features from remote sensing, climate, and topographic sources, which are available on GEE.

#### 3.2.1 Satellite optical remote sensing

The Landsat family of sensors was selected, which provide 30-meter spatial resolution global observations over decadal time spans. All available Landsat Collection 2 Tier 1 Land Surface Reflectance data from 2017 to 2024 were used, including Landsat 7 Enhanced Thematic Mapper Plus (ETM+), Landsat 8 Optical Land Imager (OLI), and Landsat 9 OLI. We extracted the reflectance data from six bands, including Red, Green, Blue, Near Infrared (NIR), Shortwave Infrared 1 (SWIR1), and Shortwave Infrared 2 (SWIR2).

#### 3.2.2 Climate reanalysis and topographic data

Climate features are widely used in agricultural modeling as they capture environmental drivers of agroecosystem dynamics and complement remote sensing features. We chose the ERA5-Land Climate Reanalysis dataset (Muñoz-Sabater et al., 2021), which provides daily global data at 10 km resolution by integrating land surface modeling with in-situ observations worldwide. From this dataset, we extracted precipitation (PPT), maximum temperature, and minimum temperature. They are used to calculate the accumulative PPT and the growing degree days (GDD) in each month of the growing season:

$$\text{PPT}_j = \sum_{d=1}^{M} \text{PPT}_d \tag{1}$$

$$\text{GDD}_j = \sum_{d=1}^{M} \sum_{h=1}^{24} (\max(0, \min(T_h - T_{min}, T_{max} - T_{min}))) \tag{2}$$

in which M denotes the number of days in the month j, $\text{PPT}_d$ is the daily total precipitation (in mm) (Eq. (1)), $T_h$ is the hourly temperature (in °C) and calculated based on sinusoidal interpolation between the daily minimum and maximum temperatures (Eq. (2)), and $T_{max}$ ($T_{min}$) denotes the upper (lower) biological temperature threshold, beyond which crop development no longer respond to temperature variations, which is 30 °C (8 °C) for soybean and 26 °C (0 °C) for winter wheat (Ritchie, 1991; Swan et al., 1987).

Moreover, topographic characteristics play a significant role in agricultural decision-making and have been widely utilized in tillage classification. We extracted Elevation data from the USGS 3D Elevation Program 10m National Map (Archuleta et al., 2017) to describe the topographic characteristics in the study area.

### 3.3 Data cleaning and downloading

We used GEE to collect and download each dataset. Before downloading the data, noisy pixels in satellite imagery were first removed by applying the per-pixel quality mask (clouds, cloud shadows, snow/ice, and water). Following that, we used the 30-meter U.S. cropland layer (CDL) (Boryan et al., 2011) to keep pixels on specific crop types and mask out non-cropland pixels. Next, observations on the remaining pixels were aggregated to the county (field) level by calculating the mean values within each county (field) boundary. Finally, the aggregated observations were downloaded from GEE, including AEF embeddings, time-series satellite remote sensing data, monthly climate features, and Elevation.

### 3.4 Remote sensing data processing

The frequency of time-series remote sensing observations varies spatially due to different satellite path overlap among adjacent tracks and cloud coverage. Therefore, it is challenging to use raw time-series observations as predictors. Instead, we pre-processed the time-series remote sensing data and generated predictor variables tailored to each task.

#### 3.4.1 Features for yield prediction

In yield prediction tasks, we downloaded the time-series satellite imagery across the whole growing season for each crop. In addition to six spectral bands, two vegetation indices (VIs), Normalized Difference Vegetation Index (NDVI) and Green Chlorophyll Vegetation Index (GCVI), were calculated (Eq. (3)-(4)). NDVI is a classic remote sensing index that measures vegetation vigor and greenness by contrasting NIR and Red reflectance. GCVI extends this approach by incorporating green reflectance, making it more directly sensitive to chlorophyll content and photosynthetic capacity while partially alleviating NDVI's saturation issues in high-biomass regions. (Gitelson et al., 2003).

$$\text{NDVI} = \frac{\text{NIR} - \text{Red}}{\text{NIR} + \text{RED}} \tag{3}$$

$$\text{GCVI} = \frac{\text{NIR}}{\text{Green}} \tag{4}$$

We fit a second-order harmonic regression (Wilson et al., 2018) to each spectral bands and VIs based on all available observations during the growing season (Eq. (5)).

$$y(t) = c + a_1 \cos(2\pi t) + b_1 \sin(2\pi t) + a_2 \cos(4\pi t) + b_2 \sin(4\pi t) \tag{5}$$

where t represents the date of the observation; $c$ denotes the intercept coefficient; $a_1$ and $b_1$ ($a_2$ and $b_2$) represent the first-order (second-order) cosine and sine coefficients, respectively.

The five harmonic coefficients ($a_1$, $b_1$, $a_2$, $b_2$, $c$) for spectral bands and VIs, which summarize the time-series satellite data, were used as predictor variables. In addition, we extracted phenology-based metrics from the fitted harmonic curves. For each spectral and VI bands, the maximum value in the growing season was picked and termed as Band_peak. In addition, band values were derived for the periods 30 days before and after the peak date, referred to as Band_b30 and Band_a30. Moreover, the partial integral approach was used to compute the area under the curve between these three time points (Deines et al., 2021). The area under the curve between Band_peak and Band_b30 was termed as Band_b30_int. The area under the curve between Band_peak and VI_a30 was termed as VI_a30_int. These indicators capture vegetation status during the phases leading up to and following the peak, offering information on growth trajectories. An example of the time-series NDVI and the derived harmonic features is present in Figure 2.

In total, 80 harmonic features were generated for each county or field, comprising 10 features for each of the 6 spectral bands and 2 VIs. Meanwhile, monthly GDD and PPT were used as predictor variables in the U.S., which added 10 additional predictors for corn and soybean (May to Sep) and 12 additional predictors for winter wheat (Jan to Jun).

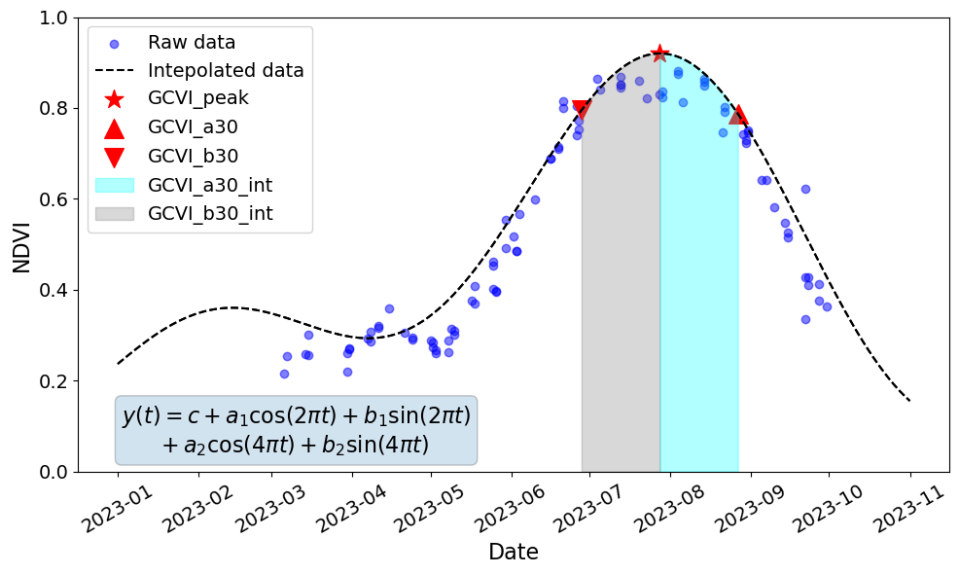

Figure 2 An example of the 2023 time-series NDVI and the derived harmonic features for corn lands in Adams County, Illinois, USA.

### 3.4.2 Features for tillage mapping

For tillage mapping, we further calculated three tillage indices that quantify crop residue cover, including NDTI, Simple Tillage Index (STI), and Crop Residue Cover index (CRC) (Eq. (6)–(8)) (Eskandari et al., 2016; Sullivan et al., 2006; Van Deventer et al., 1997):

$$\text{NDTI} = \frac{\text{SWIR1} - \text{SWIR2}}{\text{SWIR1} + \text{SWIR2}} \tag{6}$$

$$\text{STI} = \frac{\text{SWIR1}}{\text{SWIR2}} \tag{7}$$

$$\text{CRC} = \frac{\text{SWIR1} - \text{Blue}}{\text{SWIR1} + \text{Blue}} \tag{8}$$

The tillage mapping experiments were conducted in the U.S. Corn Belt, where the cropping systems are mainly corn and soybean, and most tillage is implemented in spring (Lu et al., 2022). We organized monthly remote sensing features by extracting the monthly minimum and maximum values from times-series observations in April to June for each band.

A total of 66 monthly remote sensing features was used for each county or field, comprising 2 features for each of the 6 raw bands, 2 VIs (GCVI and NDVI), and 3 tillage indices (NDTI, STI, CRC) in each month. In addition, elevation was included as a predictor variable to describe the terrain characteristics.

### 3.4.3 Features for cover crop mapping

Cover crops can be planted and terminated at any point after the harvest and before planting in the next season, a period that spans from October to May of the next year in the U.S. Corn Belt. For each field in the Corteva cover crop dataset, we collected the time-series Landsat six raw bands and two VIs (NDVI and GCVI) during this period and fitted a second-order harmonic regression. Following that, we extracted the minimum and maximum values for each band from the fitted curves within each month from October to May. Meanwhile, we calculated monthly average temperature and accumulative precipitation for each field. The extracted predictors were paired with the corresponding cover crop labels (Class 0: no cover crop; Class 1: cover crop) for model training and evaluation. The total number of predictors is 144 for each field, including 18 monthly features across 8 months.

### 3.5 Evaluation models and schemes

Two widely used ML models for tabular data were trained and evaluated for each task, including RF and XGBoosting (XGB) (Breiman, 2001). Both models were implemented using Scikit-learn with 200 individual trees (Pedregosa et al., 2011). We did not select DL models because the limited number of training samples in certain tasks would likely be insufficient to train reliable DL models. Moreover, RF and XGB have shown to outperform, or perform comparably to DL models on tabular data, while requiring substantially less computational resources (Hollmann et al., 2025). In addition, we also trained baseline models based on traditional remote sensing feature sets. Each downstream task required different predictor variables (Section 3.4), which we collectively refer to as remote sensing (RS) features. Each RF and XGB model was trained separately using AEF embeddings and RS features for each task (Table 1).

A standard scheme to evaluate ML model performance is random cross validation (CV), in which the dataset is randomly split to a training set and a test set. In geospatial tasks, random CV can lead to inflated model performance due to spatial autocorrelation, as nearby counties/fields with similar properties may be split between the training and testing sets. Instead, we evaluated the model performance in each task using two schemes: spatial-temporal CV and yearly CV. In spatial-temporal CV, samples from the same region and years are grouped into either the training or the testing set, avoiding spatial autocorrelation. Specifically, we implemented State-Year CV for county-level data, in which county samples from different state-year groups are used for training and testing, separately. Similarly, for the field-level data, we implemented County-Year CV. In addition, yearly CV reflects a more practical scenario, when data from a single year serve as the test set and data from the remaining years are used for training.

In addition, we designed scale-transfer and space-transfer experiments to assess whether embeddings enhance accuracy and generalizability across scale and spatial domains. In scale-transfer experiments, ML models for yield prediction are trained using county-level yield data and applied to predict yield at the field level. In space-transfer experiments, ML models are trained and tested across two ecoregions in the U.S. Corn Belt. Specifically, the study counties were grouped into two ecoregions as defined by the United States Environmental Protection Agency (EPA): the Eastern Temperate Forests (ETF) and the Great Plains (GP). The ETF is characterized by a warm, humid, temperate climate, with humid summers and mild to cold winters. In contrast, the GP consists mainly of flat grasslands with limited forest cover, featuring very hot summers and harsh winters (Ma et al., 2021). The ETF mainly includes five eastern states (Figure 1 (a)), including Illinois, Indiana, Michigan, Ohio, and Wisconsin. The GP mainly covers seven states on the western side of the Corn Belt (Figure 1 (a)), including Iowa, Kansas, Minesota, Missouri, North Dakota, Nebraska, and South Dakota. For simplicity, we termed ETF and GP as East and West, respectively.

Table 2 An overview of the evaluation schemes in this study.

| Scheme | Training Set | Test Set | Objective |
| --- | --- | --- | --- |
| State (County)-Year CV | Data in specific state (county)-year groups | Data in the rest state(county)-year groups | Avoid spatial and temporal autocorrelation |
| Yearly CV | Data in all but one years | Data in the remaining year | A practical scenario where training data are available in certain years |
| Scale-Transfer | County data | Field data | Evaluate the scale transferability |
| Space-Transfer | Data in one ecoregion | Data in the other ecoregion | Evaluate the spatial transferability |

We have summarized the evaluation scheme used in this study in Table 2. Each experiment is repeated five times under different random seeds, and the mean evaluation results are presented. For regression tasks, coefficient of determination ($R^2$) and root mean squared error (RMSE) are calculated as the evaluation metrics (Eq. (9) – (10)). For classification tasks, the evaluation metrics include the overall accuracy (Accuracy), F1 scores for each class (F1-0 and F1-1), and the average F1 score weighted by the number of true instances (F1-weighted) (Eq. (11) – (14)):

$$R^2 = 1 - \frac{\sum_{i=1}^{N}(y_i - \hat{y}_i)^2}{\sum_{i=1}^{N}(y_i - \bar{y})^2} \tag{9}$$

$$\text{RMSE} = \sqrt{\frac{1}{N}\sum_{i=1}^{N}(y_i - \hat{y}_i)^2} \tag{10}$$

$$\text{Accuracy} = \frac{TP + TN}{N} \tag{11}$$

$$\text{Precision} = \frac{TP}{TP + FP} \tag{12}$$

$$\text{Recall} = \frac{TP}{TP + FN} \tag{13}$$

$$F1 = 2 \cdot \frac{\text{Precision} \cdot \text{Recall}}{\text{Precision} + \text{Recall}} \tag{14}$$

where $N$ denotes the total number of samples in the evaluation set; $y_i$ is the ground truth data and $\bar{y}$ is the average value of the ground truth data; $\hat{y}_i$ is the predicted values by the models. *TP*, *TN*, and *FP* represent the number of True Positive, True Negative, and False Positive classifications by the models.

## 4 Experiment Results

### 4.1. Crop yield prediction

#### 4.1.1 County-level evaluation

In the U.S., the AEF-based models had comparable performance as the RS-based models under the State-Year CV, achieving an $R^2$ of around 0.80 for corn and soybean, and 0.70 for winter wheat (Table 3). All models showed a slight reduction in accuracy under the yearly CV scheme, yet AEF-based models consistently outperformed RS-based models, highlighting the stronger temporal generalizability of AEF embeddings.

Table 3 Evaluation results of county-level crop yield prediction in the U.S. in 2017-2024. The best performer is highlighted in bold for each case.

| Crop | Scheme | AEF | | | | RS | | | |
|---|---|---|---|---|---|---|---|---|---|
| | | RF | | XGB | | RF | | XGB | |
| | | $R^2$ | RMSE | $R^2$ | RMSE | $R^2$ | RMSE | $R^2$ | RMSE |
| Corn | State-Year | 0.78 | 1.19 | 0.78 | 1.18 | 0.79 | 1.14 | **0.80** | **1.12** |
| | Yearly | **0.77** | **1.15** | 0.77 | 1.16 | 0.76 | 1.20 | 0.74 | 1.25 |
| Soybean | State-Year | 0.75 | 0.38 | 0.78 | 0.36 | 0.78 | 0.37 | **0.79** | **0.36** |
| | Yearly | 0.74 | 0.39 | **0.77** | **0.36** | 0.71 | 0.41 | 0.70 | 0.42 |
| Winter Wheat | State-Year | 0.76 | 0.68 | **0.78** | **0.64** | 0.70 | 0.78 | 0.71 | 0.76 |
| | Yearly | 0.77 | 0.68 | **0.78** | **0.67** | 0.69 | 0.79 | 0.69 | 0.78 |

Scatter plots of the ground truth and predicted yields (Figure 3) indicate that both AEF-based and RS-based models align well with the reported yields from USDA-NASS. Considering that the XGB models slightly outperform the RF models in most cases, the comparison is presented using the results from the XGB models. RS-based models showed notable underestimation of winter wheat yields in high-yield ranges (Figure 3(c2)), primarily due to the limited number of high-yield samples in the training set. AEF-based models partially mitigated biases in high-yield ranges and achieved higher accuracy (Figure 3(c1)).

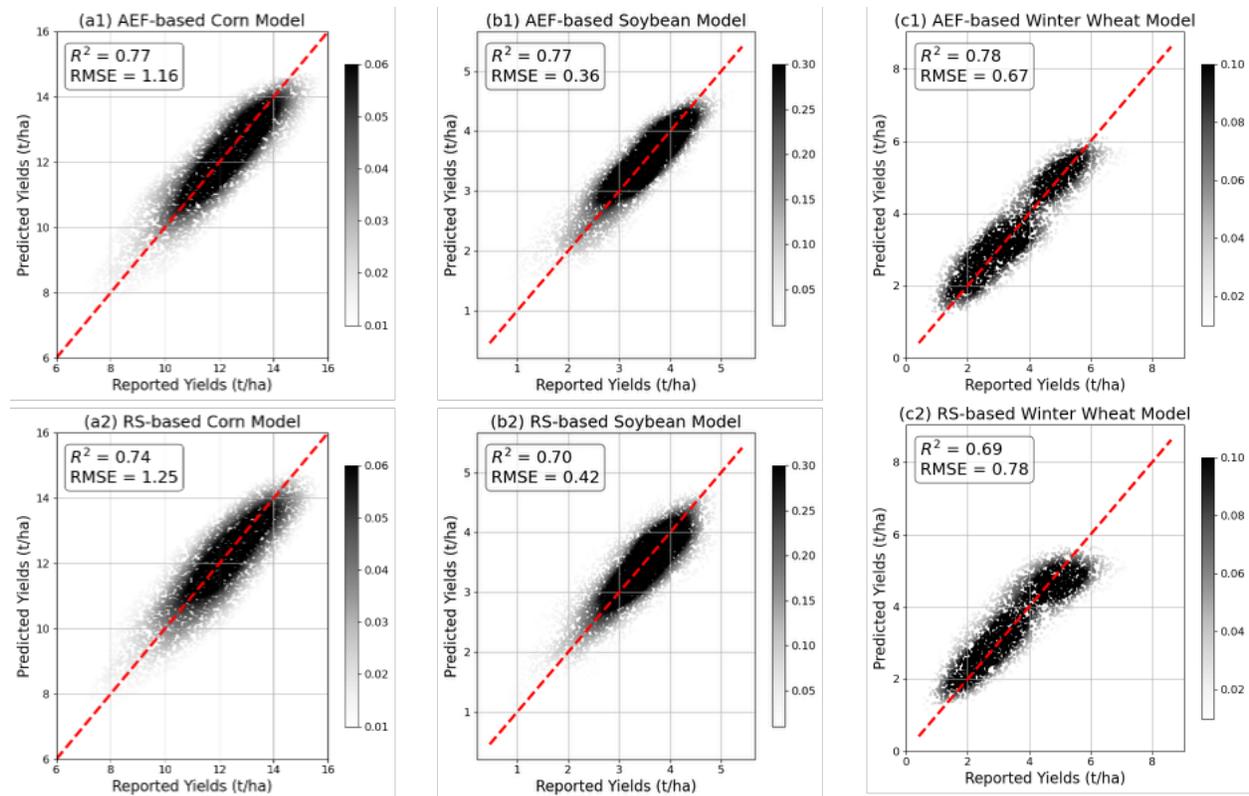

Figure 3 The scatter plots of yield prediction results by XGB under the yearly CV in the U.S. for (a) Corn, (b) Soybean, and (c) Winter Wheat in 2017-2024.

**4.1.2 Field-level evaluation**

The performance of AEF embeddings was further evaluated for field-level yield prediction based on the Corteva dataset of over 180,000 yield records (Table 1). Predicting field-level yields is typically more challenging than at the county level, primarily due to noisier yield records, reduced satellite data coverage, and the mismatch between fine-scale fields and the coarse spatial resolution of climate variables. Therefore, we observed smaller $R^2$ and larger RMSE in the field-level evaluation results (Table 4). Similar to the county-level results, AEF-based models achieved comparable performance as the RS-based models under the county-year CV and the yearly CV (Table 4).

We further compared the model performance under the scale-transfer scheme, in which the models were trained using predictors and yield records at the county level and implemented to predict yields at the field level (Ma et al., 2024b). This scheme evaluates predictors' transferability across spatial scales. AEF-based models significantly underperformed the RS-based models in this task (Table 4). This is likely due to the compression inherent in embedding representations, which, when aggregated to the county level, reduces the model's sensitivity to subtle yield-driving factors at the field scale. For instance, each single embedding may capture different signals across locations within a county. By averaging all AEF pixels to generate county-level representations, field-level information relevant to crop growth can be diluted or lost, leading to weaker predictive power at the field level. In contrast, remote sensing features preserve more direct physical signals (e.g., canopy structure, vegetation vigor, and soil moisture) that are universally interpretable across scales, enabling stronger generalization across scales.

Table 4 Evaluation results of field-level crop yield prediction in the U.S. The best performer is highlighted in bold for each case.

| Crop | Scheme | AEF | | | | RS | | | |
|---|---|---|---|---|---|---|---|---|---|
| | | RF | | XGB | | RF | | XGB | |
| | | $R^2$ | RMSE | $R^2$ | RMSE | $R^2$ | RMSE | $R^2$ | RMSE |
| Corn | County-Year | 0.43 | 2.28 | 0.44 | 2.27 | 0.45 | 2.24 | **0.46** | **2.24** |
| | Yearly | 0.31 | 2.54 | 0.30 | 2.55 | **0.33** | 2.50 | **0.33** | **2.49** |
| | Scale-Transfer | 0.24 | 2.67 | 0.25 | 2.64 | 0.37 | 2.43 | **0.38** | **2.40** |
| Soybean | County-Year | 0.36 | 0.87 | **0.37** | **0.87** | 0.35 | 0.90 | 0.35 | 0.90 |
| | Yearly | 0.29 | 0.92 | **0.31** | **0.91** | 0.28 | 0.93 | 0.25 | 0.95 |
| | Scale-Transfer | 0.13 | 1.02 | 0.11 | 1.03 | 0.22 | 0.97 | **0.24** | **0.95** |
| Winter Wheat | County-Year | **0.43** | 1.32 | **0.43** | **1.31** | 0.42 | 1.30 | 0.41 | 1.31 |
| | Yearly | 0.38 | 1.34 | **0.40** | **1.32** | 0.36 | 1.35 | 0.36 | 1.36 |
| | Scale-Transfer | -0.18 | 1.85 | -0.04 | 1.73 | 0.20 | 1.52 | **0.22** | **1.50** |

**4.2 Tillage classification**

**4.2.1 County-level tillage mapping**

For county-level tillage mapping, ML models were trained to estimate the ratio of cropland under low-intensity tillage in each county. When evaluated under the state-year CV, AEF-based and RS-based models provided broadly comparable predictive skill, with R² values around 0.48 and RMSE near 12.20% (Table 5). When assessed under yearly CV, the AEF-based RF model achieved the strongest performance in 2017 (R² = 0.56, RMSE = 11.85%), whereas for 2022 the advantage shifted toward RS-based RF model (R² = 0.53, RMSE = 12.16%). When aggregating across all years, these differences tend to smooth out, resulting in modestly higher R² for AEF-based models (0.52) compared to RS-based models (0.50), though overall accuracies remain similar (Table 5).

Table 5 Evaluation results of county-level low-intensity tillage mapping in the U.S. The best performer is highlighted in bold for each case.

| Scheme | Year | AEF | | | | RS | | | |
|---|---|---|---|---|---|---|---|---|---|
| | | RF | | XGB | | RF | | XGB | |
| | | $R^2$ | RMSE | $R^2$ | RMSE | $R^2$ | RMSE | $R^2$ | RMSE |
| State-Year | 2017&2022 | 0.47 | 12.22% | 0.47 | 12.22% | **0.48** | **12.18%** | 0.47 | 12.32% |
| Yearly | 2017 | **0.56** | **11.85%** | 0.54 | 12.04% | 0.46 | 13.00% | 0.48 | 12.79% |
| | 2022 | 0.47 | 12.91% | 0.42 | 13.59% | **0.53** | **12.16%** | 0.48 | 12.84% |
| | All | **0.52** | **12.40%** | 0.48 | 12.85% | 0.50 | 12.61% | 0.48 | 12.82% |

**4.2.2 Field-level tillage mapping**

For field-level tillage mapping, ML models were trained to classify the tillage activity of each field into high-intensity tillage (class 0) or low-intensity tillage (class 1). Under the county-year CV, RS-based models achieved slightly better performance than AEF-based models, while all models had comparable predictive power, with accuracy and weighted F1 scores around 0.75–0.80 (Figure 4). High-intensity tillage (class 0) is more difficult to detect because it requires remote sensing observations during the brief period when crop residues are removed and major field disturbances occur. Consequently, RS-based models yielded lower F1-0 scores compared to F1-1, and AEF-based models exhibited the same pattern (Figure 4).

In particular, large variations were observed in the yearly CV since the timing and intensity of tillage activities vary from year to year. RS-based models attained accuracy and weighted F1 scores around 0.75, while the scores of AEF-based models dropped to around 0.70 (Figure 5). AEF-based models had lower F1-0 scores with considerable variability. These results indicate that AEF embeddings are not inherently superior to RF features for detecting changes that occur within short time windows.

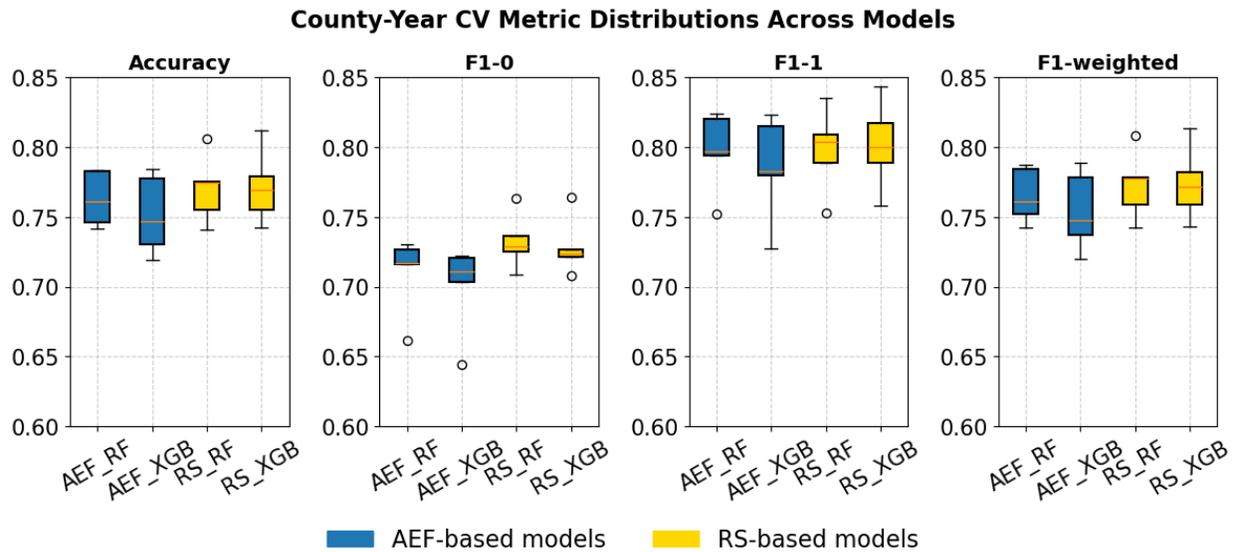

Figure 4 The field-level tillage mapping model performance (accuracy, F1-0, F1-1, and F1-weighted) under the county-year CV scheme. Each experiment is repeated five times, and each box contains the scores of all iterations.

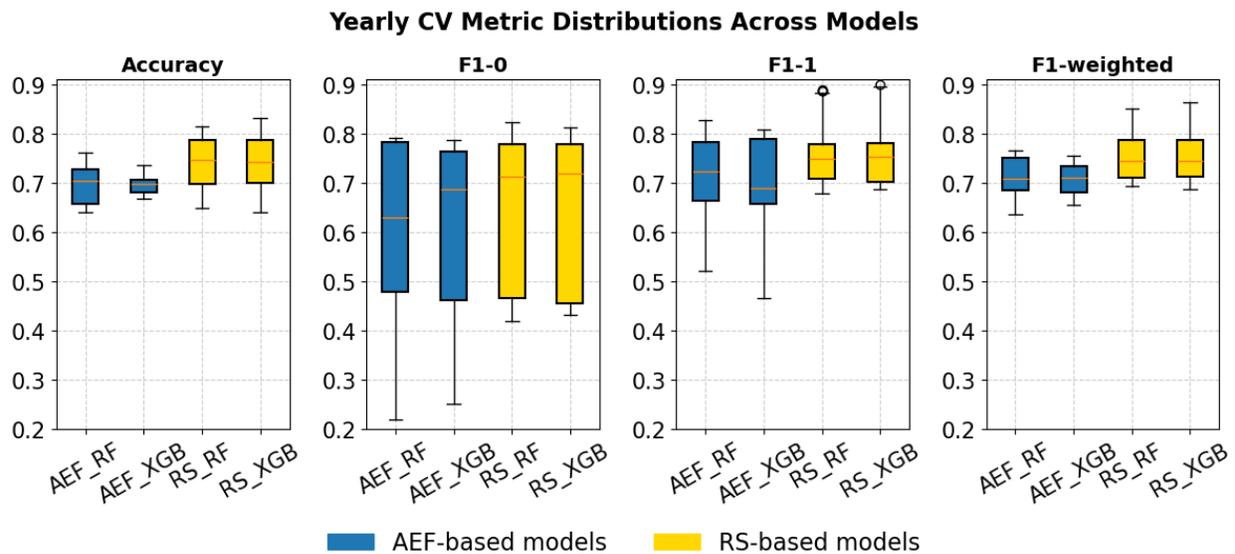

Figure 5 The field-level tillage mapping model performance (accuracy, F1-0, F1-1, and F1-weighted) under the yearly CV scheme. Each experiment is repeated five times, and each box contains the scores of all iterations.

### 4.3 Cover crop mapping

For field-level cover crop mapping, ML models were trained to classify fields as either without (class 0) or with cover crops (class 1). Generally, RF models had better performance than XGB models. The degraded performance by XGB is potentially due to the noise in the input feature sets, since there were fewer satellite observations during winter and early spring because of high cloud coverage and the presence of snow and ice. (Fig. S1), leading to more noise in the remote sensing predictors. Prior work has similarly found that RF is more robust to noisy observations than XGB (Fawagreh et al., 2014).

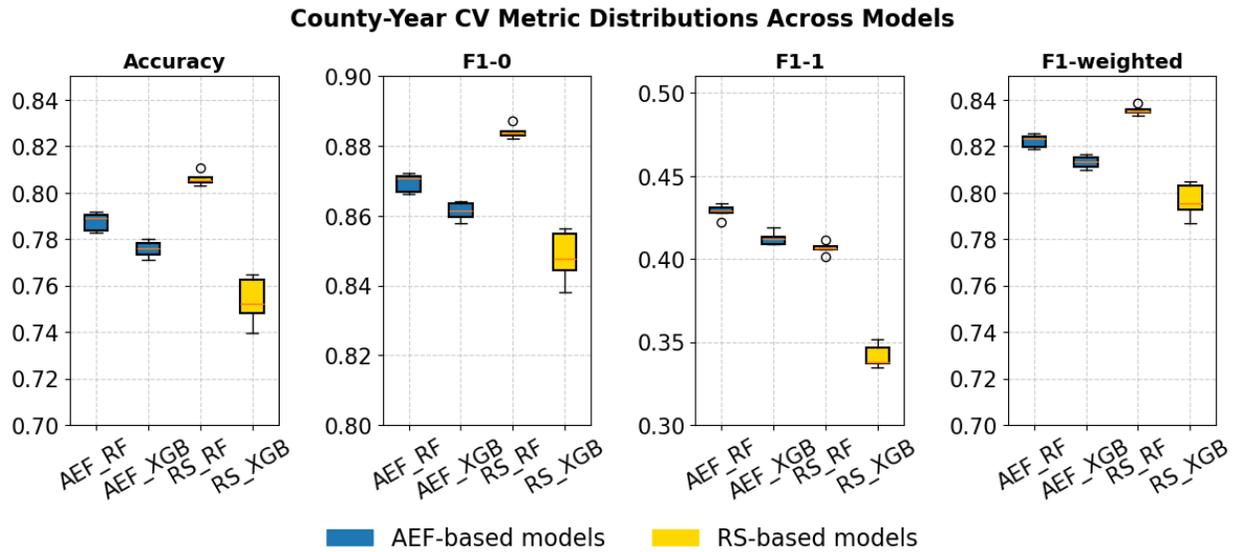

Figure 6 The field-level cover crop mapping model performance (accuracy, F1-0, F1-1, and F1-weighted) under the county-year CV scheme. Each experiment is repeated five times, and each box contains the scores of all iterations.

Under both schemes, RS-based RF models had the highest overall performance, with accuracy and weighted F1 scores around 0.80–0.85 (Figure 6–7). AEF-based models performed well in detecting cover crops (higher F1-1) but misclassified more negative samples. Again, under the yearly CV (Figure 7), all models showed large variations in performance, likely reflecting the varying timing of cover cropping from year to year. AEF-based models exhibited the largest variations (Figure 7), reflecting the disproportional representation by AEF embeddings across time.

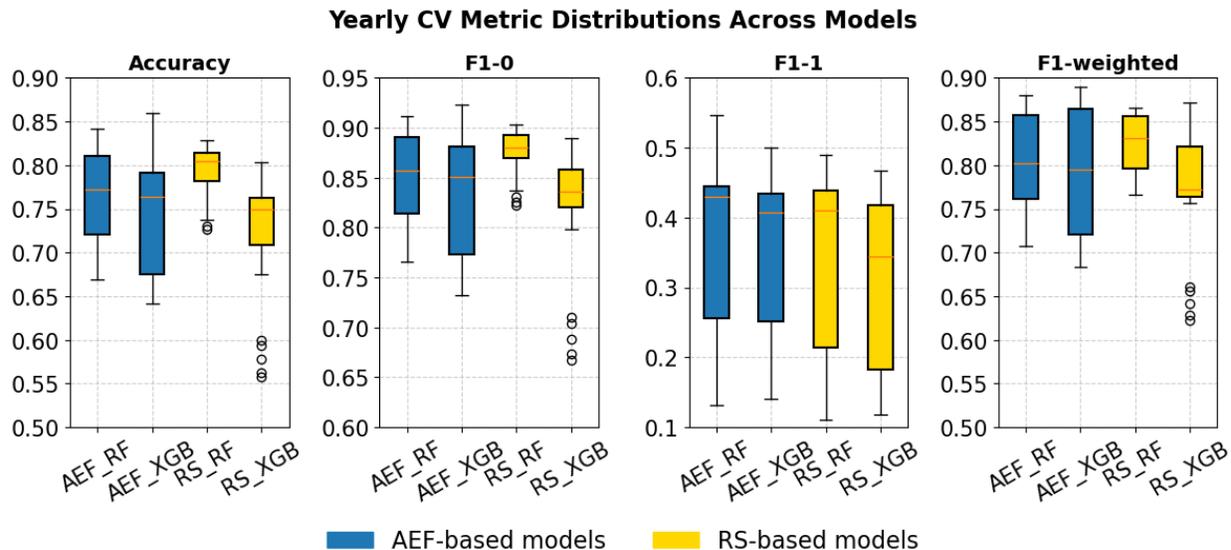

Figure 7 The field-level cover crop mapping model performance (accuracy, F1-0, F1-1, and F1-weighted) under the yearly CV scheme. Each experiment is repeated five times, and each box contains the scores of all iterations.

**4.4 Space-transfer evaluation**

Remote sensing spectral features and climate variables generally exhibit significant geographic shifts due to variations in factors such as crop planting and harvesting dates and soil types across different regions (Ma et al., 2024a; Tong and Wang, 2025). AEF was trained with more than 3 billion images worldwide, with the potential to learn spatial-invariant features. We assessed the spatial transferability of AEF embeddings for each of the downstream tasks across an eastern ecoregion ETF (East) and a western ecoregion GP (West) within the U.S. Corn Belt at both the county and the field levels. AEF-based ML models were trained to predict yields for corn and soybean in one ecoregion and then were tested in the other, and vice versa (Section 3.5). We also trained and tested the RS-based models for comparison. Winter wheat yield prediction tasks were excluded since the wheat data were primarily located in the East. In the yield prediction experiments, we only present the results by RF models since RF and XGB had very similar performance. In the field-level tillage and cover crop mapping experiments, we also only present the RF model's results since it has proven more robust in the previous experiments.

Table 6 Space-transfer evaluation results between the East and the West ecoregions in the U.S. Corn Belt for each agricultural downstream task. The best performer is highlighted in bold for each case.

| Task | Scale | Crop | Scheme | AEF-based RF | | RS-based RF | |
|---|---|---|---|---|---|---|---|
| | | | | $R^2$ | RMSE | $R^2$ | RMSE |
| Yield Prediction | County | Corn | East → West | 0.02 | 2.71 | **0.66** | **1.61** |
| | | | West → East | 0.51 | 1.23 | **0.60** | **1.12** |
| | | Soybean | East → West | -0.28 | 0.94 | **0.53** | **0.56** |
| | | | West → East | 0.39 | 0.44 | **0.59** | **0.37** |
| | Field | Corn | East → West | 0.25 | 2.65 | **0.38** | **2.41** |
| | | | West → East | 0.33 | 2.48 | **0.34** | **2.45** |
| | | Soybean | East → West | 0.08 | 0.99 | **0.19** | **0.93** |
| | | | West → East | 0.23 | 1.00 | **0.28** | **0.97** |

Under Review

|  |  |  |  | R² | RMSE | R² | RMSE |
|---|---|---|---|---|---|---|---|
| Tillage Mapping | County | N/A | East → West | -0.35 | 0.21 | **0.11** | **0.17** |
|  |  |  | West → East | **0.43** | **0.14** | 0.21 | 0.15 |
|  |  |  |  | Accuracy | F1-weighted | Accuracy | F1-weighted |
|  | Field | N/A | East → West | 0.64 | 0.64 | **0.67** | **0.68** |
|  |  |  | West → East | 0.66 | 0.67 | **0.72** | **0.72** |
|  |  |  |  | Accuracy | F1-weighted | Accuracy | F1-weighted |
| Cover Crop Mapping | Field | N/A | East → West | **0.88** | **0.87** | 0.84 | 0.85 |
|  |  |  | West → East | 0.48 | 0.53 | **0.52** | **0.57** |

In the yield prediction tasks, the AEF-based models had substantially worse space-transfer performance than the RS-based models. In particular, when trained in the East and tested in the West (East → West), AEF-based models failed to make accurate yield prediction for either corn or soybean in the West, with R² values around -0.28–0.07 (Table 6). When trained in the West and tested in the East (West → East), the AEF-based models had improved performance in the East, with R² values around 0.39–0.51, possibly due to having more states and thus more representative training samples in the West (Table 6). Meanwhile, RS-based models had stable performance under both space-transfer schemes and consistently outperformed AEF-based models (Table 6). Similar patterns were observed at the field level (Table 6).

We observed mixed patterns in the tillage mapping and cover crop mapping tasks (Table 6). In particular, in county-level tillage mapping, AEF-based models performed well when transferring from West to East, but underperformed RS-based models when transferring from East to West. At the field level, RS-based RF achieved higher accuracy under both schemes but with small margins. Similarly, in field-level cover crop mapping, AEF-based RF and RS-based RF exhibited very similar spatial transferability. One possible reason is that agricultural practices are less influenced by differences among ecoregions than crop yields.

## 5 Discussion
### 5.1 Advantages of AEF embeddings

Our experiments demonstrated several advantages of AEF embeddings over commonly used RS features:

**Automatic data harmonization**: AEF embeddings inherently harmonize data from multiple sources, including optical, multispectral, radar, LiDAR, gravity fields, climate, topographic, and text sources. AEF reconciles multiple sparse, non-uniformly observation records with varying resolutions and formats into a continuous record with a 10-meter spatial resolution. Ablation experiments in the AEF paper showed that each modality has contributed positively and increased the accuracy in downstream tasks (Figure 4 in (Brown et al., 2025)). As such, users do not need to collect data from each individual source or design feature engineering strategies to harmonize them.

**Fast data downloading and processing**: The annual AEF embeddings are stored as an image dataset on GEE, and each year's data are organized as a raster image. This data format allows fast data downloading from GEE. Also, the AEF embeddings require no pre-processing and are ready for modeling. In contrast, downloading RS imagery from GEE takes much more time, as it involves extensive data filtering, cloud masking, and spatial aggregation on each observation date. The downloaded time-series RS images require intensive pre-processing and quality control before they can be used for model training and inference.

**Competitive performance in local experiments**: AEF-based models generally are competitive with RS-based models, especially in yield prediction. For example, under the state-year CV for crop yield prediction, AEF-based XGB model achieved the $R^2$ values of 0.78 for corn, soybean, and winter wheat, compared to 0.80, 0.79, and 0.71 for the RS-based XGB model.

**Better spatial coverage**: We also noticed that AEF embedding dataset has better spatial coverage than commonly used RS imagery, as it provides consistent, gap-free representations across space and time and is not subject to cloud contamination. For instance, there are a total number of 90,210 corn yield records in the Corteva yield dataset. We successfully downloaded AEF embeddings for all but downloaded RS features for 89,938 of them. The missing RS records are mostly due to data gap in Landsat or insufficient number of time-series observations for harmonic fitting.

## 5.2 Limitations of AEF embeddings
### 5.2.1 Limited spatial transferability

AEF embeddings exhibit more pronounced geographic shifts than remote sensing features in certain tasks. Our space-transfer experiments showed that AEF-based had competitive spatial transferability in tillage and cover crop mapping but consistently underperformed RS-based models in yield prediction (Table 6). We visualized the distributions of the crop-specific AEF embeddings in each ecoregion using t-SNE, a ML-based data visualization tool that projects high-dimensional space to a 2D space (Maaten and Hinton, 2008). The distributions of AEF embeddings in each ecoregion differs markedly from each other (Figure 8), reflecting substantial geographic shifts in the feature space. The overlapping samples are limited and correspond to counties situated along the boundary between the two ecoregions. Consequently, AEF-based models trained in one region learn region-specific association between the predictors and the yield data, which can hinder transfer to the other region.

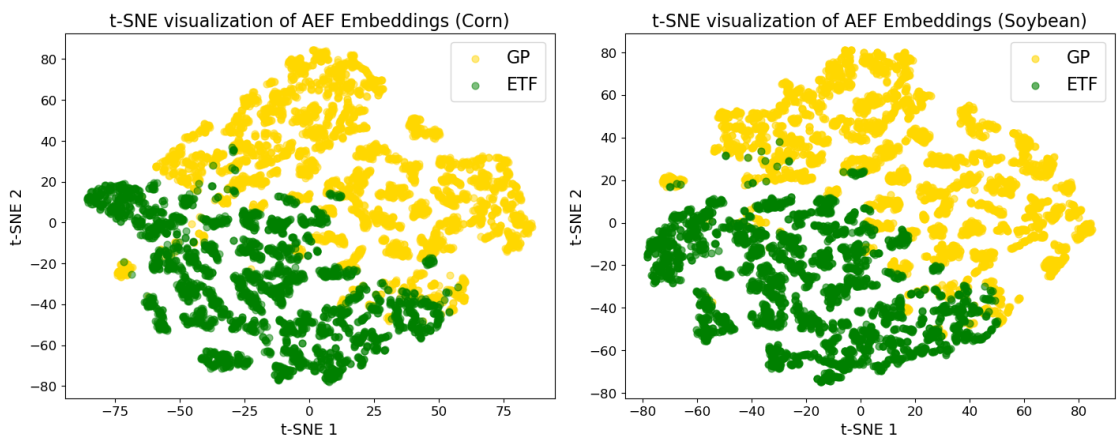

Figure 8 The t-SNE plots of the crop-specific AEF embeddings at the county level in the Eastern Temperate Forests (ETF) and the Great Plains (GP) in the U.S. Corn Belt.

These geographic shifts can be even more pronounced across countries or continents. To confirm this, we further conducted space-transfer experiments from the U.S. to Argentina for county-level soybean yield prediction in 2019–2024. Notably, AEF-based models trained in the U.S. significantly underperformed the RS-based models and failed to make accurate yield predictions in Argentina across all years (Table 7).

Table 7 Space-transfer evaluation results of county-level soybean yield prediction in Argentina.

| Scheme | Year | AEF | | | | RS | | | |
| --- | --- | --- | --- | --- | --- | --- | --- | --- | --- |
| | | RF | | XGB | | RF | | XGB | |
| | | $R^2$ | RMSE | $R^2$ | RMSE | $R^2$ | RMSE | $R^2$ | RMSE |
| US → Argentia | 2019 | -5.24 | 1.84 | -5.67 | 1.62 | 0.04 | 0.72 | **0.05** | **0.72** |
| | 2020 | -3.09 | 1.54 | -3.53 | 1.43 | -0.16 | 0.81 | **-0.14** | **0.81** |
| | 2021 | -3.04 | 1.35 | -3.54 | 1.40 | **-0.29** | **0.75** | -0.30 | 0.76 |
| | 2022 | -3.08 | 1.33 | -3.58 | 0.67 | **0.01** | **0.65** | -0.02 | 0.66 |
| | 2023 | -0.64 | 0.64 | -0.82 | 1.59 | **-0.04** | **0.53** | -0.06 | 0.53 |
| | 2024 | -3.84 | 1.51 | -4.35 | 1.49 | **0.30** | **0.58** | 0.27 | 0.59 |
| | All | -2.13 | 1.42 | -2.45 | 1.90 | **0.27** | **0.68** | 0.26 | 0.69 |

The limited spatial transferability may arise from multiple contributing factors. During model training, AEF embeddings were decoded to not only reconstruct satellite images but also DEM and gravity fields, which are region-specific and relatively static, and may limit the embeddings to capturing localized information. Another possible reason is that geolocated articles from Wikipedia were used to provide text-based information as inputs to a text encoder, which were updated via contrastive learning with vision-based models. Such text-based information might be region-specific, resulting in distinct embeddings across space, particularly when comparing regions across countries or continents.

### 5.2.2 Low interpretability

Another limitation when implementing AEF embeddings is the low interpretability. Each individual band of the 64-dimentional annual embeddings is labeled sequentially from A00 to A63, without clear physical meanings. An example of feature importance analysis is presented in Figure 9, in which the feature importance in FM-based and RS-based RF models was quantified by the mean decrease in impurity across all trees in the forest (Breiman, 2001). Based on feature importance identified by RS-based RF models (Figure 9 (b)), it is evident that satellite-derived features contribute more than climate variables, with GCVI proving more informative than NDVI and raw spectral bands. This information can guide feature selection and model design. Meanwhile, A05 is the most important feature in FM-based RF models and surpasses the second A17 by a large margin (Figure 9 (a)). However, it is uncertain why A05 is the top contributor to corn yield prediction, given that its meaning is unknown. We note that the most important features for soybean yields differed greatly from those for corn, even though the RS-based models had similar feature importance for the two crops.

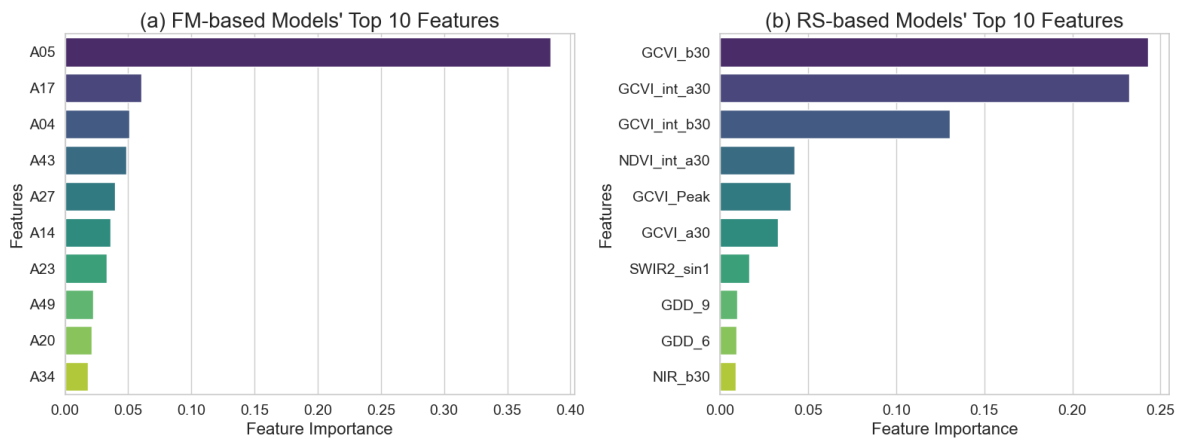

Figure 9 The Top 10 important features in (a) FM-based RF models and (b) RS-based RF models for county-level corn yield predication.

### 5.2.3 Other limitations

In addition, we observed several other limitations in AEF embeddings. One is that AEF-based models showed temporal variations under yearly CV similar to RS-based models. This finding suggests that although AEF was trained to realize time-continuous featurization, it is still subject to the availability and quality of EO, leading to featurization with varying performance across different years. Another limitation is that AEF lacks temporal sensitivity and therefore cannot support time-critical tasks such as in-season yield prediction. The annual AEF embeddings are derived from year-long time series of Earth observations, which constrains their applicability for real-time analyses.

## 6 Conclusion

The rapid development of GFMs shows promises to overcome the long-standing limitations in EO. Google DeepMind's AEF is the first GFM that supports multi-modal EO featurization across continuous time, and the global, annual, and analysis-ready AEF embeddings are available from 2017 to 2024 on GEE. Internal evaluations show that AEF embeddings outperform existing featurization approaches in multiple EO tasks. However, those tasks are mainly about land cover and land use classification, with no systematic evaluation on its performance in agricultural tasks. This study comprehensively evaluated AEF embeddings on three critical agricultural downstream tasks at both regional and field levels in the U.S., including crop yield prediction, tillage mapping, and cover crop mapping. Evaluation results show that AEF-based models have strong predictive power and are competitive with RS-based models in several scenarios. AEF also has better spatial coverage and requires minimal pre-processing before modeling. It has the potential to transform the current EO-based agricultural research.

However, we also identified several limitations in AEF, including limited spatial transferability and low interpretability. In particular, AEF-based models exhibited significant geographic shifts and failed to generate reasonable yield predictions across regions and countries. We also found it is challenging to interpret feature importance in AEF-based models, since there is no clear physical meaning in each band of the embeddings. Beyond that, other limitations in AEF embeddings include temporal variability and limited time sensitivity. Collectively, these limitations raise concerns when applying AEF embeddings in agriculture, where time sensitivity, generalizability, and interpretability is critical. Future AEF versions or other GFM development can tackle these limitations through generating monthly/seasonal embeddings, balancing training site selection, and incorporating feature attribution techniques to enhance interpretability.

**Acknowledgements**

We appreciate Sang-Zi Liang from Corteva for helping to collect and organize the field yield data. Funding was provided by NASA Acres (NASA Applied Sciences Grant No. 80NSSC23M0034, sub-award 124245-Z6512205 to D.B.L.). Funding was also provided by Google's Satellite Embeddings Dataset Small Grants Program, awarded to Y.M. and D.B.L. Any opinions, findings, and conclusions or recommendations expressed in this study are those of the authors and do not necessarily reflect the views of NASA or Google.

**Reference**

Abdalla, M., Hastings, A., Cheng, K., Yue, Q., Chadwick, D., Espenberg, M., Truu, J., Rees, R.M., Smith, P., 2019. A critical review of the impacts of cover crops on nitrogen leaching, net greenhouse gas balance and crop productivity. Global Change Biology 25, 2530–2543. https://doi.org/10.1111/gcb.14644

Alonso-Ayuso, M., Gabriel, J.L., García-González, I., Del Monte, J.P., Quemada, M., 2018. Weed density and diversity in a long-term cover crop experiment background. Crop Protection 112, 103–111. https://doi.org/10.1016/j.cropro.2018.04.012

Archuleta, C.-A.M., Constance, E.W., Arundel, S.T., Lowe, A.J., Mantey, K.S., Phillips, L.A., 2017. The National Map seamless digital elevation model specifications. US Geological Survey.

Becker-Reshef, I., Barker, B., Humber, M., Puricelli, E., Sanchez, A., Sahajpal, R., McGaughey, K., Justice, C., Baruth, B., Wu, B., Prakash, A., Abdolreza, A., Jarvis, I., 2019. The GEOGLAM crop monitor for AMIS: Assessing crop conditions in the context of global markets. Global Food Security 23, 173–181. https://doi.org/10.1016/j.gfs.2019.04.010

Boryan, C., Yang, Z., Mueller, R., Craig, M., 2011. Monitoring US agriculture: the US Department of Agriculture, National Agricultural Statistics Service, Cropland Data Layer Program. Geocarto International 26, 341–358. https://doi.org/10.1080/10106049.2011.562309

Breiman, L., 2001. Random Forests. Machine Learning 45, 5–32. https://doi.org/10.1023/A:1010933404324

Brown, C.F., Kazmierski, M.R., Pasquarella, V.J., Rucklidge, W.J., Zhang, C., Shelhamer, E., Lahera, E., Wiles, O., Ilyushchenko, S., Zhang, L.L., Alj, S., Schechter, E., Askay, S., Guinan, O., Moore, R., Boukouvalas, A., Kohli, P., 2025. AlphaEarth Foundations: An embedding field model for accurate and efficient global mapping from sparse label data.

Claassen, R., Langpap, C., Wu, J., 2017. Impacts of Federal Crop Insurance on Land Use and Environmental Quality. American Journal of Agricultural Economics 99, 592–613. https://doi.org/10.1093/ajae/aaw075

Cong, Y., Khanna, S., Meng, C., Liu, P., Rozi, E., He, Y., Burke, M., Lobell, D., Ermon, S., 2022. SatMAE: Pre-training Transformers for Temporal and Multi-Spectral Satellite Imagery. Advances in Neural Information Processing Systems 35, 197–211.

Deering, D.W., 1978. RANGELAND REFLECTANCE CHARACTERISTICS MEASURED BY AIRCRAFT AND SPACECRAFTSENSORS. Texas A&M University.

Deines, J.M., Patel, R., Liang, S.-Z., Dado, W., Lobell, D.B., 2021. A million kernels of truth: Insights into scalable satellite maize yield mapping and yield gap analysis from an extensive ground dataset in the US Corn Belt. Remote Sensing of Environment 253, 112174. https://doi.org/10.1016/j.rse.2020.112174

Eskandari, I., Navid, H., Rangzan, K., 2016. Evaluating spectral indices for determining conservation and conventional tillage systems in a vetch-wheat rotation. International Soil and Water Conservation Research 4, 93–98. https://doi.org/10.1016/j.iswcr.2016.04.002

Fawagreh, K., Gaber, M.M., Elyan, E., 2014. Random forests: from early developments to recent advancements. Systems Science & Control Engineering 2, 602–609. https://doi.org/10.1080/21642583.2014.956265

Fendrich, A.N., Matthews, F., Van Eynde, E., Carozzi, M., Li, Z., d'Andrimont, R., Lugato, E., Martin, P., Ciais, P., Panagos, P., 2023. From regional to parcel scale: A high-resolution map of cover crops across Europe combining satellite data with statistical surveys. Science of The Total Environment 873, 162300. https://doi.org/10.1016/j.scitotenv.2023.162300

Gitelson, A.A., Viña, A., Arkebauer, T.J., Rundquist, D.C., Keydan, G., Leavitt, B., 2003. Remote estimation of leaf area index and green leaf biomass in maize canopies. Geophysical research letters 30.

Guan, K., Jin, Z., Peng, B., Tang, J., DeLucia, E.H., West, P.C., Jiang, C., Wang, S., Kim, T., Zhou, W., Griffis, T., Liu, L., Yang, W.H., Qin, Z., Yang, Q., Margenot, A., Stuchiner, E.R., Kumar, V., Bernacchi, C., Coppess, J., Novick, K.A., Gerber, J., Jahn, M., Khanna, M., Lee, D., Chen, Z., Yang, S.-J., 2023. A scalable framework for quantifying field-level agricultural carbon outcomes. Earth-Science Reviews 243, 104462. https://doi.org/10.1016/j.earscirev.2023.104462

He, K., Chen, X., Xie, S., Li, Y., Dollar, P., Girshick, R., 2022. Masked Autoencoders Are Scalable Vision Learners, in: 2022 IEEE/CVF Conference on Computer Vision and Pattern Recognition (CVPR). Presented at the 2022 IEEE/CVF Conference on Computer Vision and Pattern Recognition (CVPR), IEEE, New Orleans, LA, USA, pp. 15979–15988. https://doi.org/10.1109/CVPR52688.2022.01553

Hollmann, N., Müller, S., Purucker, L., Krishnakumar, A., Körfer, M., Hoo, S.B., Schirrmeister, R.T., Hutter, F., 2025. Accurate predictions on small data with a tabular foundation model. Nature 637, 319–326. https://doi.org/10.1038/s41586-024-08328-6

Kauth, R.J., Thomas, G.S., 1976. The tasselled cap--a graphic description of the spectral-temporal development of agricultural crops as seen by Landsat, in: LARS Symposia. p. 159.

Kogan, F., Kussul, N., Adamenko, T., Skakun, S., Kravchenko, O., Kryvobok, O., Shelestov, A., Kolotii, A., Kussul, O., Lavrenyuk, A., 2013. Winter wheat yield forecasting in Ukraine based on Earth observation, meteorological data and biophysical models. International Journal of Applied Earth Observation and Geoinformation 23, 192–203. https://doi.org/10.1016/j.jag.2013.01.002

Koudahe, K., Allen, S.C., Djaman, K., 2022. Critical review of the impact of cover crops on soil properties. International Soil and Water Conservation Research 10, 343–354. https://doi.org/10.1016/j.iswcr.2022.03.003

Lobell, D.B., Di Tommaso, S., Zhou, Q., Ma, Y., Specht, J., Guan, K., 2025. The mixed effects of recent cover crop adoption on US cropland productivity. Nat Sustain 1–9. https://doi.org/10.1038/s41893-025-01599-5

Lu, C., Yu, Z., Hennessy, D.A., Feng, H., Tian, H., Hui, D., 2022. Emerging weed resistance increases tillage intensity and greenhouse gas emissions in the US corn–soybean cropping system. Nat Food 3, 266–274. https://doi.org/10.1038/s43016-022-00488-w

Luo, D., Zhang, H.K., Houborg, R., Ndekelu, L.M.N., Maimaitijiang, M., Tran, K.H., McMaine, J., 2023. Utility of daily 3 m Planet Fusion Surface Reflectance data for tillage practice mapping with deep learning. Science of Remote Sensing 7, 100085. https://doi.org/10.1016/j.srs.2023.100085

Ma, Y., Chen, S., Ermon, S., Lobell, D.B., 2024a. Transfer learning in environmental remote sensing. Remote Sensing of Environment 301, 113924. https://doi.org/10.1016/j.rse.2023.113924

Ma, Y., Liang, S.-Z., Myers, D.B., Swatantran, A., Lobell, D.B., 2024b. Subfield-level crop yield mapping without ground truth data: A scale transfer framework. Remote Sensing of Environment 315, 114427. https://doi.org/10.1016/j.rse.2024.114427

Ma, Y., Zhang, Z., Yang, H.L., Yang, Z., 2021. An adaptive adversarial domain adaptation approach for corn yield prediction. Computers and Electronics in Agriculture 187, 106314.

Maaten, L. van der, Hinton, G., 2008. Visualizing data using t-SNE. Journal of machine learning research 9, 2579–2605.

Muñoz-Sabater, J., Dutra, E., Agustí-Panareda, A., Albergel, C., Arduini, G., Balsamo, G., Boussetta, S., Choulga, M., Harrigan, S., Hersbach, H., Martens, B., Miralles, D.G., Piles, M., Rodríguez-Fernández, N.J., Zsoter, E., Buontempo, C., Thépaut, J.-N., 2021. ERA5-Land: a state-of-the-art global reanalysis dataset for land applications. Earth System Science Data 13, 4349–4383. https://doi.org/10.5194/essd-13-4349-2021

Nakalembe, C., Becker-Reshef, I., Bonifacio, R., Hu, G., Humber, M.L., Justice, C.J., Keniston, J., Mwangi, K., Rembold, F., Shukla, S., Urbano, F., Whitcraft, A.K., Li, Y., Zappacosta, M., Jarvis, I., Sanchez, A., 2021. A review of satellite-based global agricultural monitoring systems available for Africa. Global Food Security 29, 100543. https://doi.org/10.1016/j.gfs.2021.100543

Pedregosa, F., Varoquaux, G., Gramfort, A., Michel, V., Thirion, B., Grisel, O., Blondel, M., Prettenhofer, P., Weiss, R., Dubourg, V., Vanderplas, J., Passos, A., Cournapeau, D., Brucher, M., Perrot, M., Duchesnay, É., 2011. Scikit-learn: Machine Learning in Python. Journal of Machine Learning Research 12, 2825–2830.

Plastina, A., Liu, F., Miguez, F., Carlson, S., 2020. Cover crops use in Midwestern US agriculture: perceived benefits and net returns. Renewable Agriculture and Food Systems 35, 38–48. https://doi.org/10.1017/S1742170518000194

Reed, C.J., Gupta, R., Li, S., Brockman, S., Funk, C., Clipp, B., Keutzer, K., Candido, S., Uyttendaele, M., Darrell, T., 2023. Scale-MAE: A Scale-Aware Masked Autoencoder for Multiscale Geospatial Representation Learning, in: 2023 IEEE/CVF International Conference on Computer Vision (ICCV). Presented at the 2023 IEEE/CVF International Conference on Computer Vision (ICCV), IEEE, Paris, France, pp. 4065–4076. https://doi.org/10.1109/ICCV51070.2023.00378

Ritchie, J.T., 1991. Wheat Phasic Development, in: Modeling Plant and Soil Systems. John Wiley & Sons, Ltd, pp. 31–54. https://doi.org/10.2134/agronmonogr31.c3

Sullivan, D.G., Truman, C.C., Schomberg, H.H., Endale, D.M., Strickland, T.C., 2006. Evaluating techniques for determining tillage regime in the Southeastern Coastal Plain and Piedmont. Agronomy journal 98, 1236–1246.

Swan, J.B., Schneider, E.C., Moncrief, J.F., Paulson, W.H., Peterson, A.E., 1987. Estimating Corn Growth, Yield, and Grain Moisture from Air Growing Degree Days and Residue Cover1. Agronomy Journal 79, 53–60. https://doi.org/10.2134/agronj1987.00021962007900010012x

Szwarcman, D., Roy, S., Fraccaro, P., Gíslason, Þ.E., Blumenstiel, B., Ghosal, R., Oliveira, P.H. de, Almeida, J.L. de S., Sedona, R., Kang, Y., Chakraborty, S., Wang, S., Kumar, A., Truong, M., Godwin, D., Lee, H., Hsu, C.-Y., Asanjan, A.A., Mujeci, B., Keenan, T., Arevalo, P., Li, W., Alemohammad, H., Olofsson, P., Hain, C., Kennedy, R., Zadrozny, B., Cavallaro, G., Watson, C., Maskey, M., Ramachandran, R., Moreno, J.B., 2024. Prithvi-EO-2.0: A Versatile Multi-Temporal Foundation Model for Earth Observation Applications. https://doi.org/10.48550/arXiv.2412.02732

Tong, X.-Y., Wang, S., 2025. Invariant Features for Global Crop Type Classification. https://doi.org/10.48550/arXiv.2509.03497

USDA-NASS, 2024. Census of Agriculture Methodology. USDA.

Van Deventer, A.P., Ward, A.D., Gowda, P.H., Lyon, J.G., 1997. Using thematic mapper data to identify contrasting soil plains and tillage practices. Photogrammetric engineering and remote sensing 63, 87–93.

Wang, Y., Albrecht, C.M., Braham, N.A.A., Mou, L., Zhu, X.X., 2022. Self-Supervised Learning in Remote Sensing: A review. IEEE Geoscience and Remote Sensing Magazine 10, 213–247. https://doi.org/10.1109/MGRS.2022.3198244

Wilson, B.T., Knight, J.F., McRoberts, R.E., 2018. Harmonic regression of Landsat time series for modeling attributes from national forest inventory data. ISPRS Journal of Photogrammetry and Remote Sensing 137, 29–46. https://doi.org/10.1016/j.isprsjprs.2018.01.006

Wu, K., Zhang, Yingying, Ru, L., Dang, B., Lao, J., Yu, L., Luo, J., Zhu, Z., Sun, Y., Zhang, J., Zhu, Q., Wang, J., Yang, M., Chen, J., Zhang, Yongjun, Li, Y., 2025. A semantic-enhanced multi-modal remote sensing foundation model for Earth observation. Nat Mach Intell 7, 1235–1249. https://doi.org/10.1038/s42256-025-01078-8

Xiong, X., Zhong, R., Jiang, H., Athanasiadis, I., Yang, Y., Zhu, L., Lin, T., 2026. Corn yield estimation under extreme climate stress with knowledge-encoded deep learning. ISPRS Journal of Photogrammetry and Remote Sensing 231, 101–118. https://doi.org/10.1016/j.isprsjprs.2025.10.020

Xiong, Z., Wang, Y., Zhang, F., Stewart, A.J., Hanna, J., Borth, D., Papoutsis, I., Saux, B.L., Camps-Valls, G., Zhu, X.X., 2024. Neural Plasticity-Inspired Multimodal Foundation Model for Earth Observation. https://doi.org/10.48550/arXiv.2403.15356

Zhang, C., Kerner, H., Wang, S., Hao, P., Li, Z., Hunt, K.A., Abernethy, J., Zhao, H., Gao, F., Di, L., Guo, C., Liu, Z., Yang, Z., Mueller, R., Boryan, C., Chen, Q., Beeson, P.C., Zhang, H.K., Shen, Y., 2025. Remote sensing for crop mapping: A perspective on current and future crop-specific land cover data products. Remote Sensing of Environment 330, 114995. https://doi.org/10.1016/j.rse.2025.114995

Zhou, Q., Guan, K., Wang, Sheng, Jiang, C., Huang, Y., Peng, B., Chen, Z., Wang, Sibo, Hipple, J., Schaefer, D., Qin, Z., Stroebel, S., Coppess, J., Khanna, M., Cai, Y., 2022. Recent Rapid Increase of Cover Crop Adoption Across the U.S. Midwest Detected by Fusing Multi-Source Satellite Data. Geophysical Research Letters 49, e2022GL100249. https://doi.org/10.1029/2022GL100249

Zulauf, C., Schnitkey, G., Paulson, N., Coppess,  and J., 2024. Cover Crops and Covered Cropland, 2022 US Census of Agriculture. farmdoc daily 14.

**Appendix**:

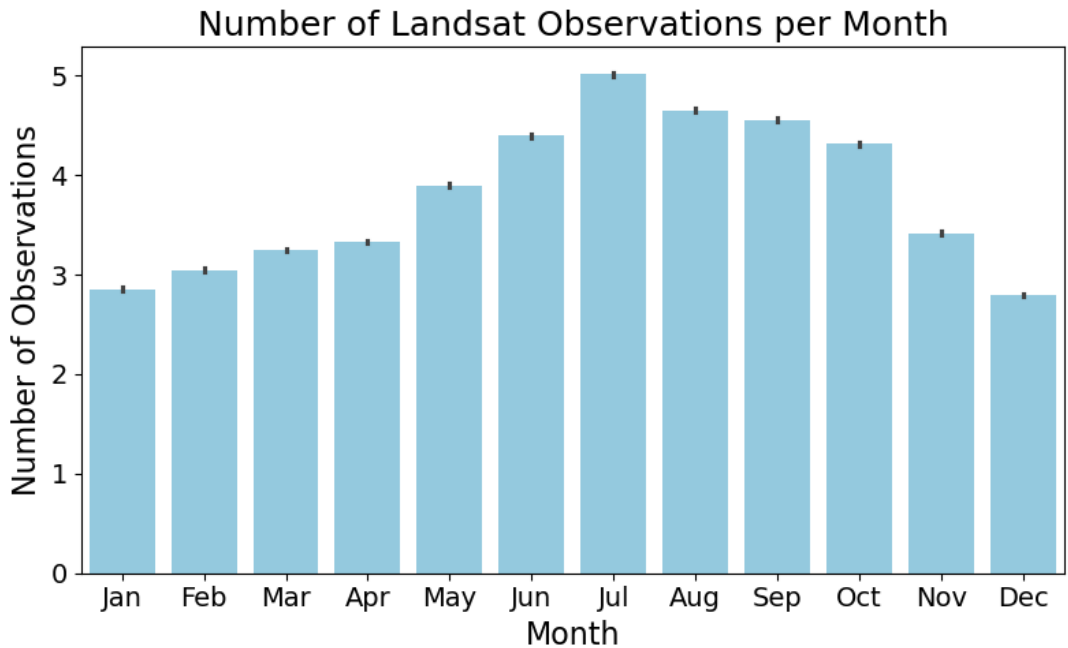

Figure S1. The average number of Landsat observation per month for fields in the Corteva cover crop dataset. There are fewer observations during winter and early spring, leading to more noise in the feature set for cover crop mapping.